\documentclass[
]{ceurart}

\usepackage{listings}
\usepackage{hyperref}
\usepackage{amssymb}
\usepackage{pifont}
\newcommand{\cmark}{\ding{51}}%
\newcommand{\xmark}{\ding{55}}%
\usepackage{caption}
\usepackage{subcaption}
\usepackage{tabularx}

\begin{document}

\copyrightyear{2022}
\copyrightclause{Copyright for this paper by its authors.
  Use permitted under Creative Commons License Attribution 4.0
  International (CC BY 4.0).}

\conference{IRCDL 2023 – XIX: The Conference on Information and Research science Connecting to Digital and Library science, 23-24 February 2023 - Bari, Italy}

\title{CTE: A Dataset for Contextualized Table Extraction}


\author[1]{Andrea Gemelli}[%
orcid=0000-0002-6149-8282,
email=andrea.gemelli@unifi.it,
url=https://andreagemelli.github.io,
]
\cormark[1]
\fnmark[1]

\author[1]{Emanuele Vivoli}[%
orcid=0000-0002-9971-8738,
email=emanuele.vivoli@unifi.it,
url=http://www.emanuelevivoli.me,
]
\fnmark[1]

\author[1]{Simone Marinai}[%
orcid=0000-0002-6702-2277,
email=simone.marinai@unifi.it,
url=https://tinyurl.com/simone-marinai,
]
\fnmark[1]

\address[1]{Dipartimento di Ingegneria dell’Informazione (DINFO)
Università degli studi di Firenze, Italy}

\cortext[1]{Corresponding author.}
\fntext[1]{The authors contributed equally.}

\begin{abstract}
Relevant information in documents is often summarized in tables, helping the reader to identify useful facts. 
Most benchmark datasets support either document layout analysis or table understanding, but lack in providing data to apply both tasks in a unified way.
We define the task of Contextualized Table Extraction (CTE), which aims to extract and define the  structure of tables considering the textual context of the document.
The dataset comprises 75k fully annotated pages of scientific papers, including more than 35k tables.
Data are gathered from PubMed Central, merging the information provided by annotations in the  PubTables-1M and PubLayNet datasets. The dataset can support CTE and adds new classes to the original ones.
The generated annotations can be used to develop end-to-end pipelines for various tasks, including document layout analysis, table detection, structure recognition, and functional analysis.
We formally define CTE and evaluation metrics, showing which subtasks can be tackled, describing advantages, limitations, and future works of this collection of data. 
Annotations and code will be accessible at \href{https://github.com/AILab-UniFI/cte-dataset}{https://github.com/AILab-UniFI/cte-dataset}.

\end{abstract}


\maketitle

\section{Introduction}

Nowadays, large collections of documents require a huge amount of human work to annotate documents and extract important information. In the last thirty years, the community of Document Analysis and Recognition (DAR) tried to overcome this challenge, exploiting suitable algorithms and artificial intelligence techniques to automatize the analysis of documents and reduce its costs. Among others, Document Classification (DC), Layout Analysis (DLA), and Table Understanding (TU) more broadly attracted the interest of researchers and companies. 
DC is the first step of many DAR pipelines, since different kinds of documents require different strategies: given a document, either scanned or digital-born, the aim is to classify it into a specific category, e.g. invoice or magazine. 
DLA \cite{13-00-Marinai} aims at recognizing homogeneous regions within the document, grouping smaller components close to each other such as regions of text, and, if required, assigning it a category (e.g. a title or an image caption). Finally, TU \cite{21-06-TableUnderstanding} is an umbrella term for table detection and recognition: tables summarize important information within documents and their detection along with the recognition of their structure is crucial to automatically query collections of documents.

During the past years, the interest in the detection and recognition of tables raised significantly, leading to the automation of important processes such as information extraction. In particular, for scientific literature, it is crucial to extract tabular data, e.g. to make the research comparable and help scholars to reconstruct the SOTA of the different fields of study \cite{20-04-AxCell}. Moreover, collections of scientific papers such as arXiv and PubMed opened to the possibility of accessing a large number of documents along with their structural information represented in standard formats such as \LaTeX \; and XML. That is why scientific literature parsing and scientific table analysis rapidly became one of the most prominent areas of research in DAR: large datasets have been released \cite{19-08-PubLayNet, 21-10-PubTables-1M}, allowing the community to develop deep learning models. Unfortunately, as we will describe in the next sections, these datasets come with partial information that forces the experimentation of layout analysis and table extraction separately. From this identified lack, we define Contextualized Table Extraction, a broad task that comes along with novel annotations for a collection of 75k scientific pages containing  more than 35k tables, encouraging the development of new systems capable of tackling a multitude of tasks at once. 

In this paper, we introduce a new task called Contextualized Table Extraction that is a framework, which involves detecting tables, recognizing their structure, and performing functional analysis in an end-to-end manner. 
CTE is formulated as a token and link classification task, which allows for multiple tasks to be addressed simultaneously overcoming common limitations such as being performed separately or lacking a comprehensive dataset.  CTE is built on top of well known tasks in DAR. 
CTE is designed to be suitable for methods employing Graph Neural Networks, which are widely used in applications where the structure and layout in documents matter. 
We provide a new set of labels structured in a way that allows us to merge information of selected scientific publications from other well known benchmark datasets. In this way we obtain a comprehensive dataset for the task of CTE. 
We believe that the combination of methods applied to process the labeled documents and produce the merged information collected is a novel contribution to the field of document analysis as well.

\begin{table}[t]
    \centering
    \caption{Comparison of CTE with related datasets: $\clubsuit$ \ \ denotes the datasets used to generate the new annotations. A dataset is $S4G$ (suitable for graphs) if a graph can be constructed directly with no further preprocessing. DLA (Document Layout Analysis), TD (Table Detection), TSR (Table Structure Recognition), and TFA (Table Functional Analysis) show which tasks models can be trained for.}
    \resizebox{\linewidth}{!}{
    \begin{tabular}{l | ccc | cccc | c}
        Dataset  &  \#pages  &  \#tables  & \#classes  & DLA  & TD  & TSR  & TFA  & S4G \\ \midrule
        PubLayNet ($\clubsuit$) & 358k & 107k & 5 & \cmark & \cmark & \xmark & \xmark & \xmark \\
        PubTables-1M ($\clubsuit$) & 574k & 948k & 7 & \xmark & \cmark & \cmark & \cmark & \xmark \\
        DocBank & 500k & 417k* & 12 & \cmark & \cmark & \cmark* & \xmark & \cmark** \\
        SciTSR & 0 & 15k & - & \xmark & \xmark & \cmark & \xmark & \cmark \\ \midrule
        CTE & 75k & 35k 
        & 13 & \cmark & \cmark & \cmark & \cmark & \cmark \\
        \bottomrule
        \multicolumn{9}{l}{{\scriptsize *DocBank is an extension of TableBank, from which we gathered these information}} \\
        \multicolumn{9}{l}{{\scriptsize **If tokens used as graph nodes, no information on edges}}
    \end{tabular}}
    \label{tab:datasets}
\end{table}

\subsection{Related Work}
Despite the advances in the field, several challenges strongly limited the generalization of methods developed until a few years ago. In particular, we can mention: (i) data quality (e.g. scanned documents or images captured in-the-wild); (ii) contents, due to different languages and/or scripts; (iii) document layouts (which differentiate in, e.g. magazines, scientific papers, and invoices). 
To address these challenges a large number of data need to be collected in order to fully exploit the power of Deep Learning models that achieve the state-of-the-art for the aforementioned tasks.
Unfortunately, creating such datasets is nothing but trivial since  accurate annotations come at a high cost in terms of time and human effort\ \cite{siegelnECCV16figureseer, doclaynet2022}. 
On the other hand, automatic annotation techniques are not always applicable since they require 
a large number of documents shared together with their source files in standard formats such as \LaTeX, XML, or HTML\ \cite{20-12-DocBank, 19-08-PubLayNet}. Additionally, these techniques usually generate weakly labeled collections and are more error-prone than manually annotated ones.

Since online archives of scientific papers are freely and publicly available along with the corresponding source information (e.g. arXiv and PubMed) several datasets have been proposed so far in the field of scientific literature parsing. 
Among others, we summarize in Table \ref{tab:datasets} some of the most important datasets proposed for layout analysis and table extraction. 
PubLayNet and DocBank have been widely used to train object detectors \cite{15-12-FasterRCNN, 17-10-MaskRCNN} and transformers \cite{19-06-LayoutLM} for DLA. 
Overall, these datasets contain around half a million pages labeled into five and twelve different classes, respectively. 
PubLayNet has been constructed merging the information extracted from PDFMiner (bounding box regions) and the XML files shared by the publishers (containing the region labels).
DocBank is built gathering the \LaTeX source files and assigning labels taking into account the section tags. 
For the Table Extraction task, a recent dataset has been released  (PubTables-1M) which  counts nearly one million tables, labeled to perform not only TD and TSR but also Table Functional Analysis (TFA) that provides  additional information on table cells like table headers. 
Even if it is smaller, SciTSR \cite{21-10-SciTSR} introduced a collection of 15k tables generated from \LaTeX\; to perform TSR, mainly using a Graph Neural Network (GNN). 
Despite this contribution, GNNs also have the advantage of being lightweight compared to transformer-based architectures while still retaining good performance, as shown in the framework Doc2Graph \cite{gemelli2022doc2graph} for document analysis.

As it is possible to notice in Table\ \ref{tab:datasets}, all these datasets lack a comprehensive and broader set of annotations, forcing the community to develop multiple systems that, in application scenarios, would lead to heavy and large pipelines. 

\subsection{Contributions}
Our ongoing work  brings several novelties, that are discussed throughout the paper and are summarized as follows: 
\begin{itemize}
    \item We define the task of Contextualized Table Extraction, an extended version of table extraction as defined in \cite{21-10-PubTables-1M} that adds layout information and encourages the development of end-to-end systems that can tackle multiple tasks at once;
    \item Novel annotations are created by merging subset of \cite{19-08-PubLayNet,21-10-PubTables-1M} that can be found in our repo\footnote{\href{https://github.com/AILab-UniFI/cte-dataset}{https://github.com/AILab-UniFI/cte-dataset}}. Our collection comprehends 75k scientific pages and more than 35k tables. Tokens at the basis of annotations correspond to words  extracted from PDFs using PyMuPDF and labeled according to the region they belong to; table structure information is encoded as links between tokens;
    \item The dataset encourages the use and development of graph methods on documents, providing to the community a new set of labeled data to experiment with GNN-based techniques. The annotations do not require any further processing (either in labels or data themselves) to construct a graph over the scientific pages.
\end{itemize}

The paper is organized as follows: in Section 2 we describe in detail how the dataset has been created and how the annotations are presented, along with some limitations we aim to address in the near future. Section 3 formalizes the CTE task by means of token and link classification. Finally, in Section 4 and 5 we
discuss future work and draw conclusions.

\section{Dataset Description}
\label{sec:dataset-descritption}

Contextualized Table Extraction (CTE), as we describe deeply in Section \ref{sec:task-and-metrics}, involves not only detecting tables, recognizing their layout and functional structure, but also takes into consideration their surrounding information. We formalize CTE to be accomplished through token and link classification, allowing multiple tasks to be tackled at once. The F1 score for CTE is defined as the average of F1 scores for token and link classification.

Although it is easy to freely access large collections of scientific papers (i.e. from arXiv or PubMed Central) it is difficult to find documents labeled with complete information.
Most benchmark datasets support either DLA or TU. 
However, as our aim is encouraging the development of systems capable of tackling more tasks at once, a new dataset is needed.
The proposed dataset for CTE is obtained by merging data and annotations given by PubLayNet and PubTables-1M datasets, both based on PubMed Central publications. As depicted in the next sections, firstly we identify the pages of scientific papers annotated in both datasets, then we merge the information and add two novel classes (captions and page information) and finally use PyMuPDF to extract text and position of tokens. We used a preliminary small version of this collection in \cite{22-08-Gemelli}, applying a GNN to tackle CTE.
After the release of PubLayNet test set we updated the version of CTE dataset, now containing more annotated data.

\subsection{Subset of PubLayNet and PubTables-1M}

PubLayNet is a collection of $358,353$ PDF pages with five types of regions annotated (\textit{title, text, list, table, image})  \cite{19-08-PubLayNet}. 
PubTables-1M \cite{21-10-PubTables-1M} is a collection of $947,642$ fully annotated tables, including information for table detection, recognition, and functional analysis (such as identifying \textit{column headers, projected rows}, and \textit{table cells}). The datasets are built to address different tasks, as summarized in Table \ref{tab:datasets}.

To merge the datasets,  we first identify the papers belonging to both collections.
From this subset, we keep pages with tables fully annotated in PubTables-1M and pages without tables: this filters out even more pages, since we found some PubTables-1M annotations to have only one annotated table in pages containing two or more tables. Following this step, we obtain approximately 75k pages.
The resulting merged dataset contains objects labeled into 13 different classes, having in addition to the regions annotated in PubLayNet the table annotations described in PubTables-1M (\textit{row, column, table header, projected header, table cell}, and \textit{grid cell}). Moreover, we added two classes: \textit{caption} and \textit{other}. \textit{Captions} are heuristically found taking into account the proximity with images and tables, while the \textit{other} class contains all the remaining not-labeled text regions (e.g. page headers and page numbers).

The GitHub repository of our dataset is at its second version, after adding the test-set released by PubLayNet \footnote{From PubLayNet Github repo: "07/Mar/2022 - We have released the ground truth of the test set for the ICDAR 2021 Scientific Literature Parsing competition available \hyperlink{https://github.com/ibm-aur-nlp/PubLayNet/tree/master/ICDAR_SLR_competition/final_test_set}{here}."}.
We followed PubLayNet for the train/val/test splits.

\begin{figure}[t]
    \centering
    \includegraphics[width=.9\textwidth]{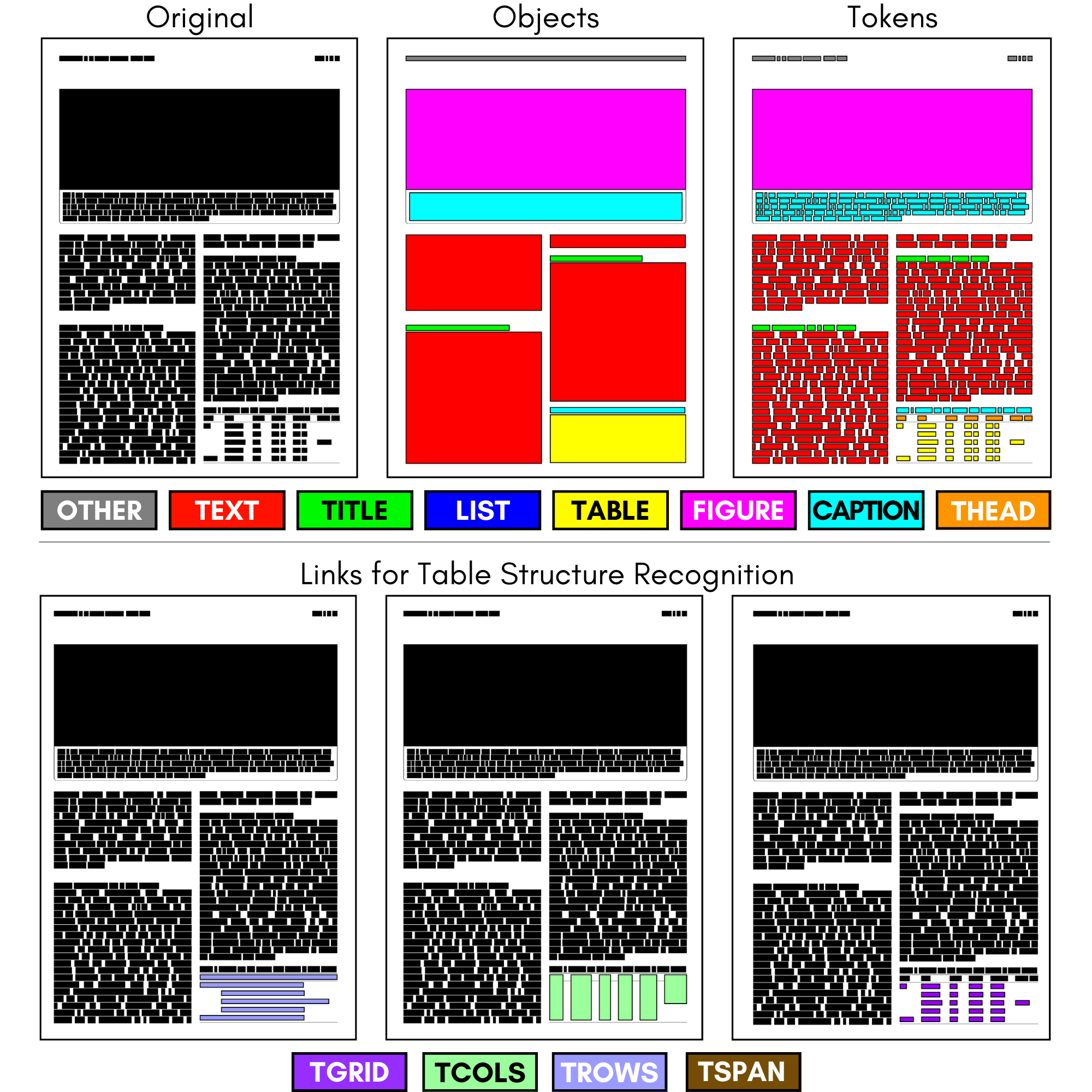}
    \caption{Example page (content is intentionally concealed in 'Original') and corresponding CTE annotations.
    Objects represent the layout regions. Tokens contain the word tokens labeled according to the class in the top of the figure. 
Acronyms for table annotations are: THEAD (table headers), TSPAN (table sub-headers spanning along different columns), TGRID (table cells), TCOLS and TROWS (respectively columns and rows of the tables).}
    \label{fig:annotaions}
\end{figure}

\subsection{Annotation procedure}
Once a complete annotated list of pages is selected from the two datasets, we leverage an external tool to extract page tokens. After comparing several tools, we opted for PyMuPDF \cite{12-00-PyMuPDF} which is a Python open-source library backed by a large community and constantly maintained. Each element, visible or not visible, present in the PDF page is extracted and annotated based on the annotation bounding-box it appears in, as depicted in Figure \ref{fig:annotaions}: tokens are labeled according to their enclosing labeled region (upper part); links, instead, are presented as groups of tokens for visualization purposes (bottom part), but encoded as couples as described in details in the next Section and in Table \ref{tab:annotations}. By doing so, the resulting page is composed by extracting page tokens along with their position (bounding boxes coordinates) and their textual content (mostly single words). 
This process heavily depends on original versions of the PDF files: even if the document name is the same along the two datasets annotations (PubLayNet and PubTables-1M) the PDF version of PubLayNet documents could differ. This is due to the two years gap between the datasets release date. To obtain reliable information, in our approach we discard all the pages (and tables) in which the content of the two sources does not correspond anymore.

\subsection{Dataset structure and format}
After the merging procedure, we end up with three JSON files (subset of the original PubLayNet one) splitting the data into train, val, and test.
Each one contains information regarding tokens extracted by PyMuPDF, their links and the regions that group them (larger objects).
Tokens have these information: {\em token id, bounding box coordinates, text, class id}, and {\em object id} (larger region to which it belongs).
Links between tokens (belonging to the same row, column or grid cell) have information such as {\em link id, class id}, and {\em token id} (list of tokens linked together). Finally, objects contain information such as {\em object id, bounding box coordinates}\ and {\em  class id}. A representation of the aforementioned annotation format is represented in Tables \ref{tab:annotations}.

\begin{table}
\centering
\caption{Annotation Format: each line contains different information in case of Objects, Tokens, or Links.}
\subcaption*{Object annotations}
\begin{tabular}{|l|c|c|c|c|c|c|}
\hline
\textbf{Index}   & \textbf{0} & \textbf{1,0} & \textbf{1,1} & \textbf{1,2} & \textbf{1,3} & \textbf{2} \\ \hline
\textbf{Content} & object id  & x0           & y0           & x1           & y1           & class id   \\ \hline
\end{tabular}

\vspace{0.2cm}
\subcaption*{Tokens annotations}
\begin{tabular}{|l|c|c|c|c|c|c|c|c|}
\hline
\textbf{Index}   & \textbf{0} & \textbf{1,0} & \textbf{1,1} & \textbf{1,2} & \textbf{1,3} & \textbf{2} & \textbf{3} & \textbf{4}       \\ \hline
\textbf{Content} & token id   & x0           & y0           & x1           & y1           & text       & class id   & parent object id \\ \hline
\end{tabular}

\vspace{0.2cm}
\subcaption*{Link annotations}
\begin{tabular}{|l|c|c|c|c|c|}
\hline
\textbf{Index}   & \textbf{0} & \textbf{1} & \textbf{2,0} & \textbf{2,1} & \textbf{2,n-1}  \\ \hline
\textbf{Content} & link id    & class id   & 1st token id & 2nd token id & n-th token id \\ \hline
\end{tabular}
\label{tab:annotations}
\end{table}

\subsection{Limitations of the Dataset}
\label{sec:limitations}
We are aware that the proposed dataset, even if it is proposing a new benchmark to tackle CTE, has room for improvement. As such, in the following we list the limitations of the dataset:
\begin{enumerate}
    \item There is a small amount of data and tables compared to  other datasets. Considering that adding more annotated data would be nothing but trivial, we believe this point could be addressed in two ways: i) as a starting pool of data to train generative models and getting new samples automatically labeled (e.g. using techniques similar to \cite{Pisaneschi-PRL2023}); ii) using the CTE collection as a challenging benchmark to compare lightweight models, such as GNNs, along with state-of-the-art transformers (notably anger of huge amount of data).
    \item The heuristics used for the the classes  {\em caption} and {\em other} could affect the generalization of trained models, highly dependent on the paper format used in PubMed Central. On the other hand, we are enriching information about tables by recognizing captions, that contain valuable table descriptions and that otherwise would be discarded.
    \item We still lack additional information such as author, keywords, and equations. We are going to add these additional labels in the near future, considering Grobid \cite{GROBID} in the annotation procedure, since it is a machine learning library for extracting technical information from scientific publications, from PDF to XML/TEI structured documents.
    \item The first attempts to define a baselines are reported in \cite{22-08-Gemelli}, in which the task of TE and DLA are treated end-to-end. This paper aims at sharing the CTE dataset in a way that the scientific community can further propose baselines on this work.
\end{enumerate}


\section{Contextualized Table Extraction}

\label{sec:task-and-metrics}

Contextualized Table Extraction (CTE) is the broader task of extracting tables (meaning their detection) recognizing their structure and performing functional analysis, along with other page layout information. 
To do so, CTE is formulated as a token and link classification tasks, similarly to \cite{20-12-DocBank}, since fine-grained objects like tokens permit to tackle multiple tasks at once.
For instance, recognizing the table headers and grid cells allows us to detect the tables (grouping tokens together through links) and add functional information. 
In addition, through token and link classification the need for more components would be reduced since a method capable of successfully solving CTE would require to train only one model, extracting more information at once.

Given Precision and Recall for token and link classification, namely Token Precision (TP), Token Recall (TR), Link Precision (LP), and Link Recall (LR). We can define the $F1_{CTE}$ metric as follows:
\begin{equation}
    F1_{CTE} = \frac{F1_{Token}+F1_{Link}}{2} =
    \frac{TP\cdot TR}{TP + TR} + \frac{LP\cdot LR}{LP + LR}.
    \label{equazione}
\end{equation}

\noindent
\textbf{Token classification}\newline
The first step required to tackle CTE is the classification of tokens, extracted from PDF pages using PyMuPDF. Tokens contain textual and positional information, along with class information inherited from the larger region they belong to (details in Table \ref{tab:annotations}, tokens annotations). This subtask exposes these properties:
\begin{enumerate}
    \item Through token classification it is possible to achieve DLA, TD, and TFA at once.
    \item If tackled along with link classification to achieve CTE the  $F1_{CTE}$ metric (Eq.\ \ref{equazione}) should be used. Instead, if tackled alone the metric proposed in \cite{20-12-DocBank} can be used as well.
\end{enumerate}

\noindent
\textbf{Link Classification}\newline
In order to group together tokens belonging to tables into columns, rows, or grid cells, additional information on links among pairs of tokens is added. This subtask exposes these properties:
\begin{enumerate}
    \item Through link classification it is possible to perform TSR.
    \item Similarly to token classification, F1 is preferred to evaluate link classification if tackled alone.
    \item Links connecting non-tables items should be considered as an additional class '{\em none}'.
\end{enumerate}

\noindent
\textbf{Object Recognition}\newline
Even if not required to do CTE, the annotations include area information of different regions in the paper (as common for object detection). Grouping together tokens belonging to the same class via edges can be exploited to find such areas, e.g. extracting sub-graphs from the whole document. A recent paper\ \cite{22-01-Google} exploited GNN to perform post-OCR paragraph recognition  by grouping together similar items in the pages. 

\subsection{Limitation of the Task}
While we acknowledge that CTE has some limitations, we believe that it represents a significant step towards a more comprehensive solution for table extraction in documents. 
In our previous work \cite{22-08-Gemelli}, we investigated different ways to achieve CTE through ablation studies, so as to analyze the impact of different components on the system's performances. In this paper, we define a metric, ($F1_{CTE}$), for the updated dataset regarding CTE. As the combination of two metrics, namely Token F1 and Link F1, they can be used to evaluate the performance of the system. 

\section{Future work}

In addition to providing a new dataset for contextualized table extraction, the CTE task can also serve as a basis for future research. One area of research is to investigate the effectiveness of using graph neural networks (GNNs) versus transformer architectures for the CTE task. The models might  be pre-trained and fine-tuned on all the original data from \cite{19-06-LayoutLM} and \cite{19-08-PubLayNet}. Comparing a lightweight network, GNN-based, with a heavy network, such as transformer-based, can help determine which approach is best suited for the CTE task. 
Another potential avenue for future work is to investigate the use of the CTE dataset for information extraction tasks, specifically in the context of scientific papers. Many papers include tables with important information that can be challenging to extract automatically, and incorporating external knowledge bases could further improve performance. With the CTE dataset, it would be possible to explore how to effectively combine table structure information with external knowledge to answer questions based on scientific papers.
Other open research questions that could be addressed using the CTE dataset include investigating cross-lingual performance, transfer learning, and developing techniques to handle different types of tables (e.g., nested tables, tables with merged cells).

\section{Conclusions}
In this work we presented a new dataset to tackle the task of Contextualized Table Extraction. The dataset is obtained by merging two well-known benchmark  datasets (PubTables-1M and PubLayNet). Usually, table extraction pipelines involve several components to perform different tasks on tables, without considering other important information present in the document such as captions. Based on these limitations, the proposed collection of data aims at developing models capable of tackling more tasks at once, resulting in CTE. Moreover, the annotations format encourages the development of systems based on GNN, that lack of a common benchmark within the DAR community for tasks different from TSR. We are looking to extend the dataset by adding more information such as authors, keywords, and equations.

\bibliography{main.bib}

\begin{thebibliography}{18}
\expandafter\ifx\csname natexlab\endcsname\relax\def\natexlab#1{#1}\fi
\providecommand{\url}[1]{\texttt{#1}}
\providecommand{\href}[2]{#2}
\providecommand{\path}[1]{#1}
\providecommand{\DOIprefix}{doi:}
\providecommand{\ArXivprefix}{arXiv:}
\providecommand{\URLprefix}{URL: }
\providecommand{\Pubmedprefix}{pmid:}
\providecommand{\doi}[1]{\href{http://dx.doi.org/#1}{\path{#1}}}
\providecommand{\Pubmed}[1]{\href{pmid:#1}{\path{#1}}}
\providecommand{\bibinfo}[2]{#2}
\ifx\xfnm\relax \def\xfnm[#1]{\unskip,\space#1}\fi
\bibitem[{Marinai(2013)}]{13-00-Marinai}
\bibinfo{author}{S.~Marinai},
\newblock \bibinfo{title}{Learning algorithms for document layout analysis},
\newblock in: \bibinfo{editor}{C.~Rao}, \bibinfo{editor}{V.~Govindaraju}
  (Eds.), \bibinfo{booktitle}{Handbook of Statistics},
  volume~\bibinfo{volume}{31} of \textit{\bibinfo{series}{Handbook of
  Statistics}}, \bibinfo{publisher}{Elsevier}, \bibinfo{address}{.},
  \bibinfo{year}{2013}, pp. \bibinfo{pages}{400--419}.
  \DOIprefix\doi{https://doi.org/10.1016/B978-0-444-53859-8.00016-3}.
\bibitem[{Hashmi et~al.(2021)Hashmi, Liwicki, Stricker, Afzal, Afzal, and
  Afzal}]{21-06-TableUnderstanding}
\bibinfo{author}{K.~A. Hashmi}, \bibinfo{author}{M.~Liwicki},
  \bibinfo{author}{D.~Stricker}, \bibinfo{author}{M.~A. Afzal},
  \bibinfo{author}{M.~A. Afzal}, \bibinfo{author}{M.~Z. Afzal},
\newblock \bibinfo{title}{Current status and performance analysis of table
  recognition in document images with deep neural networks},
\newblock \bibinfo{journal}{{IEEE} Access} \bibinfo{volume}{9}
  (\bibinfo{year}{2021}) \bibinfo{pages}{87663--87685}. \URLprefix
  \url{https://doi.org/10.1109/ACCESS.2021.3087865}.
  \DOIprefix\doi{10.1109/ACCESS.2021.3087865}.
\bibitem[{Kardas et~al.(2020)Kardas, Czapla, Stenetorp, Ruder, Riedel, Taylor,
  and Stojnic}]{20-04-AxCell}
\bibinfo{author}{M.~Kardas}, \bibinfo{author}{P.~Czapla},
  \bibinfo{author}{P.~Stenetorp}, \bibinfo{author}{S.~Ruder},
  \bibinfo{author}{S.~Riedel}, \bibinfo{author}{R.~Taylor},
  \bibinfo{author}{R.~Stojnic},
\newblock \bibinfo{title}{Axcell: Automatic extraction of results from machine
  learning papers},
\newblock in: \bibinfo{editor}{B.~Webber}, \bibinfo{editor}{T.~Cohn},
  \bibinfo{editor}{Y.~He}, \bibinfo{editor}{Y.~Liu} (Eds.),
  \bibinfo{booktitle}{Proceedings of the 2020 Conference on Empirical Methods
  in Natural Language Processing, {EMNLP} 2020, Online, November 16-20, 2020},
  \bibinfo{publisher}{Association for Computational Linguistics},
  \bibinfo{year}{2020}, pp. \bibinfo{pages}{8580--8594}. \URLprefix
  \url{https://doi.org/10.18653/v1/2020.emnlp-main.692}.
  \DOIprefix\doi{10.18653/v1/2020.emnlp-main.692}.
\bibitem[{Zhong et~al.(2019)Zhong, Tang, and Jimeno{-}Yepes}]{19-08-PubLayNet}
\bibinfo{author}{X.~Zhong}, \bibinfo{author}{J.~Tang},
  \bibinfo{author}{A.~Jimeno{-}Yepes},
\newblock \bibinfo{title}{Publaynet: Largest dataset ever for document layout
  analysis},
\newblock in: \bibinfo{booktitle}{2019 International Conference on Document
  Analysis and Recognition, {ICDAR} 2019, Sydney, Australia, September 20-25,
  2019}, \bibinfo{publisher}{{IEEE}}, \bibinfo{year}{2019}, pp.
  \bibinfo{pages}{1015--1022}. \URLprefix
  \url{https://doi.org/10.1109/ICDAR.2019.00166}.
  \DOIprefix\doi{10.1109/ICDAR.2019.00166}.
\bibitem[{Smock et~al.(2021)Smock, Pesala, and Abraham}]{21-10-PubTables-1M}
\bibinfo{author}{B.~Smock}, \bibinfo{author}{R.~Pesala},
  \bibinfo{author}{R.~Abraham},
\newblock \bibinfo{title}{{PubTables-1M}: Towards a universal dataset and
  metrics for training and evaluating table extraction models},
\newblock \bibinfo{journal}{CoRR} \bibinfo{volume}{abs/2110.00061}
  (\bibinfo{year}{2021}). \URLprefix \url{https://arxiv.org/abs/2110.00061}.
  \href{http://arxiv.org/abs/2110.00061}{{\tt arXiv:2110.00061}}.
\bibitem[{Siegel et~al.(2016)Siegel, Horvitz, Levin, Divvala, and
  Farhadi}]{siegelnECCV16figureseer}
\bibinfo{author}{N.~Siegel}, \bibinfo{author}{Z.~Horvitz},
  \bibinfo{author}{R.~Levin}, \bibinfo{author}{S.~Divvala},
  \bibinfo{author}{A.~Farhadi},
\newblock \bibinfo{title}{Figureseer: Parsing result-figures in research
  papers},
\newblock in: \bibinfo{booktitle}{European Conference on Computer Vision
  ({ECCV})}, \bibinfo{year}{2016}.
\bibitem[{Pfitzmann et~al.(2022)Pfitzmann, Auer, Dolfi, Nassar, and
  Staar}]{doclaynet2022}
\bibinfo{author}{B.~Pfitzmann}, \bibinfo{author}{C.~Auer},
  \bibinfo{author}{M.~Dolfi}, \bibinfo{author}{A.~S. Nassar},
  \bibinfo{author}{P.~W.~J. Staar},
\newblock \bibinfo{title}{Doclaynet: A large human-annotated dataset for
  document-layout analysis}  (\bibinfo{year}{2022}). \URLprefix
  \url{https://arxiv.org/abs/2206.01062}.
  \DOIprefix\doi{10.1145/3534678.353904}.
\bibitem[{Li et~al.(2020)Li, Xu, Cui, Huang, Wei, Li, and Zhou}]{20-12-DocBank}
\bibinfo{author}{M.~Li}, \bibinfo{author}{Y.~Xu}, \bibinfo{author}{L.~Cui},
  \bibinfo{author}{S.~Huang}, \bibinfo{author}{F.~Wei},
  \bibinfo{author}{Z.~Li}, \bibinfo{author}{M.~Zhou},
\newblock \bibinfo{title}{Docbank: {A} benchmark dataset for document layout
  analysis},
\newblock in: \bibinfo{editor}{D.~Scott}, \bibinfo{editor}{N.~Bel},
  \bibinfo{editor}{C.~Zong} (Eds.), \bibinfo{booktitle}{Proceedings of the 28th
  International Conference on Computational Linguistics, {COLING} 2020,
  Barcelona, Spain (Online), December 8-13, 2020},
  \bibinfo{publisher}{International Committee on Computational Linguistics},
  \bibinfo{year}{2020}, pp. \bibinfo{pages}{949--960}. \URLprefix
  \url{https://doi.org/10.18653/v1/2020.coling-main.82}.
  \DOIprefix\doi{10.18653/v1/2020.coling-main.82}.
\bibitem[{Ren et~al.(2015)Ren, He, Girshick, and Sun}]{15-12-FasterRCNN}
\bibinfo{author}{S.~Ren}, \bibinfo{author}{K.~He}, \bibinfo{author}{R.~B.
  Girshick}, \bibinfo{author}{J.~Sun},
\newblock \bibinfo{title}{Faster {R-CNN:} towards real-time object detection
  with region proposal networks},
\newblock in: \bibinfo{editor}{C.~Cortes}, \bibinfo{editor}{N.~D. Lawrence},
  \bibinfo{editor}{D.~D. Lee}, \bibinfo{editor}{M.~Sugiyama},
  \bibinfo{editor}{R.~Garnett} (Eds.), \bibinfo{booktitle}{Advances in Neural
  Information Processing Systems 28: Annual Conference on Neural Information
  Processing Systems 2015, December 7-12, 2015, Montreal, Quebec, Canada},
  \bibinfo{year}{2015}, pp. \bibinfo{pages}{91--99}. \URLprefix
  \url{https://proceedings.neurips.cc/paper/2015/hash/14bfa6bb14875e45bba028a21ed38046-Abstract.html}.
\bibitem[{He et~al.(2017)He, Gkioxari, Doll{\'{a}}r, and
  Girshick}]{17-10-MaskRCNN}
\bibinfo{author}{K.~He}, \bibinfo{author}{G.~Gkioxari},
  \bibinfo{author}{P.~Doll{\'{a}}r}, \bibinfo{author}{R.~B. Girshick},
\newblock \bibinfo{title}{Mask {R-CNN}},
\newblock in: \bibinfo{booktitle}{{IEEE} International Conference on Computer
  Vision, {ICCV} 2017, Venice, Italy, October 22-29, 2017},
  \bibinfo{publisher}{{IEEE} Computer Society}, \bibinfo{year}{2017}, pp.
  \bibinfo{pages}{2980--2988}. \URLprefix
  \url{https://doi.org/10.1109/ICCV.2017.322}.
  \DOIprefix\doi{10.1109/ICCV.2017.322}.
\bibitem[{Xu et~al.(2019)Xu, Li, Cui, Huang, Wei, and Zhou}]{19-06-LayoutLM}
\bibinfo{author}{Y.~Xu}, \bibinfo{author}{M.~Li}, \bibinfo{author}{L.~Cui},
  \bibinfo{author}{S.~Huang}, \bibinfo{author}{F.~Wei},
  \bibinfo{author}{M.~Zhou},
\newblock \bibinfo{title}{Layoutlm: Pre-training of text and layout for
  document image understanding},
\newblock \bibinfo{journal}{CoRR} \bibinfo{volume}{abs/1912.13318}
  (\bibinfo{year}{2019}). \URLprefix \url{http://arxiv.org/abs/1912.13318}.
  \href{http://arxiv.org/abs/1912.13318}{{\tt arXiv:1912.13318}}.
\bibitem[{Chi et~al.(2019)Chi, Huang, Xu, Yu, Yin, and Mao}]{21-10-SciTSR}
\bibinfo{author}{Z.~Chi}, \bibinfo{author}{H.~Huang}, \bibinfo{author}{H.~Xu},
  \bibinfo{author}{H.~Yu}, \bibinfo{author}{W.~Yin}, \bibinfo{author}{X.~Mao},
\newblock \bibinfo{title}{Complicated table structure recognition},
\newblock \bibinfo{journal}{CoRR} \bibinfo{volume}{abs/1908.04729}
  (\bibinfo{year}{2019}). \URLprefix \url{http://arxiv.org/abs/1908.04729}.
  \href{http://arxiv.org/abs/1908.04729}{{\tt arXiv:1908.04729}}.
\bibitem[{Gemelli et~al.(2022{\natexlab{a}})Gemelli, Biswas, Civitelli,
  Llad{\'o}s, and Marinai}]{gemelli2022doc2graph}
\bibinfo{author}{A.~Gemelli}, \bibinfo{author}{S.~Biswas},
  \bibinfo{author}{E.~Civitelli}, \bibinfo{author}{J.~Llad{\'o}s},
  \bibinfo{author}{S.~Marinai},
\newblock \bibinfo{title}{Doc2graph: a task agnostic document understanding
  framework based on graph neural networks},
\newblock \bibinfo{journal}{arXiv preprint arXiv:2208.11168}
  (\bibinfo{year}{2022}{\natexlab{a}}).
\bibitem[{Gemelli et~al.(2022{\natexlab{b}})Gemelli, Vivoli, and
  Marinai}]{22-08-Gemelli}
\bibinfo{author}{A.~Gemelli}, \bibinfo{author}{E.~Vivoli},
  \bibinfo{author}{S.~Marinai},
\newblock \bibinfo{title}{Graph neural networks and representation embedding
  for table extraction in {PDF} documents},
\newblock in: \bibinfo{booktitle}{26th International Conference on Pattern
  Recognition, {ICPR} 2022, Montreal, QC, Canada, August 21-25, 2022},
  \bibinfo{publisher}{{IEEE}}, \bibinfo{year}{2022}{\natexlab{b}}, pp.
  \bibinfo{pages}{1719--1726}. \URLprefix
  \url{https://doi.org/10.1109/ICPR56361.2022.9956590}.
  \DOIprefix\doi{10.1109/ICPR56361.2022.9956590}.
\bibitem[{PyMuPDF and McKie(2012)}]{12-00-PyMuPDF}
\bibinfo{author}{PyMuPDF}, \bibinfo{author}{J.~X. McKie},
  \bibinfo{title}{Pymupdf: Python bindings for mupdf's rendering library.},
  \bibinfo{howpublished}{\url{https://github.com/pymupdf/PyMuPDF}},
  \bibinfo{year}{2012}.
\bibitem[{Pisaneschi et~al.(2023)Pisaneschi, Gemelli, and
  Marinai}]{Pisaneschi-PRL2023}
\bibinfo{author}{L.~Pisaneschi}, \bibinfo{author}{A.~Gemelli},
  \bibinfo{author}{S.~Marinai},
\newblock \bibinfo{title}{Automatic generation of scientific papers for data
  augmentation in document layout analysis},
\newblock \bibinfo{journal}{Pattern Recognition Letters} \bibinfo{volume}{167}
  (\bibinfo{year}{2023}) \bibinfo{pages}{38--44}. \URLprefix
  \url{https://www.sciencedirect.com/science/article/pii/S0167865523000247}.
  \DOIprefix\doi{https://doi.org/10.1016/j.patrec.2023.01.018}.
\bibitem[{GROBID(2021)}]{GROBID}
GROBID, \bibinfo{title}{Grobid},
  \bibinfo{howpublished}{\url{https://github.com/kermitt2/grobid}},
  \bibinfo{year}{2008--2021}.
  \href{http://arxiv.org/abs/1:dir:dab86b296e3c3216e2241968f0d63b68e8209d3c}{{\tt
  arXiv:1:dir:dab86b296e3c3216e2241968f0d63b68e8209d3c}}.
\bibitem[{Wang et~al.(2022)Wang, Fujii, and Popat}]{22-01-Google}
\bibinfo{author}{R.~Wang}, \bibinfo{author}{Y.~Fujii}, \bibinfo{author}{A.~C.
  Popat},
\newblock \bibinfo{title}{Post-ocr paragraph recognition by graph convolutional
  networks},
\newblock in: \bibinfo{booktitle}{{IEEE/CVF} Winter Conference on Applications
  of Computer Vision, {WACV} 2022, Waikoloa, HI, USA, January 3-8, 2022},
  \bibinfo{publisher}{{IEEE}}, \bibinfo{year}{2022}, pp.
  \bibinfo{pages}{2533--2542}. \URLprefix
  \url{https://doi.org/10.1109/WACV51458.2022.00259}.
  \DOIprefix\doi{10.1109/WACV51458.2022.00259}.

\end{thebibliography}

\begin{figure}
\begin{subfigure}{.33\textwidth}
  \centering
  \includegraphics[width=\linewidth]{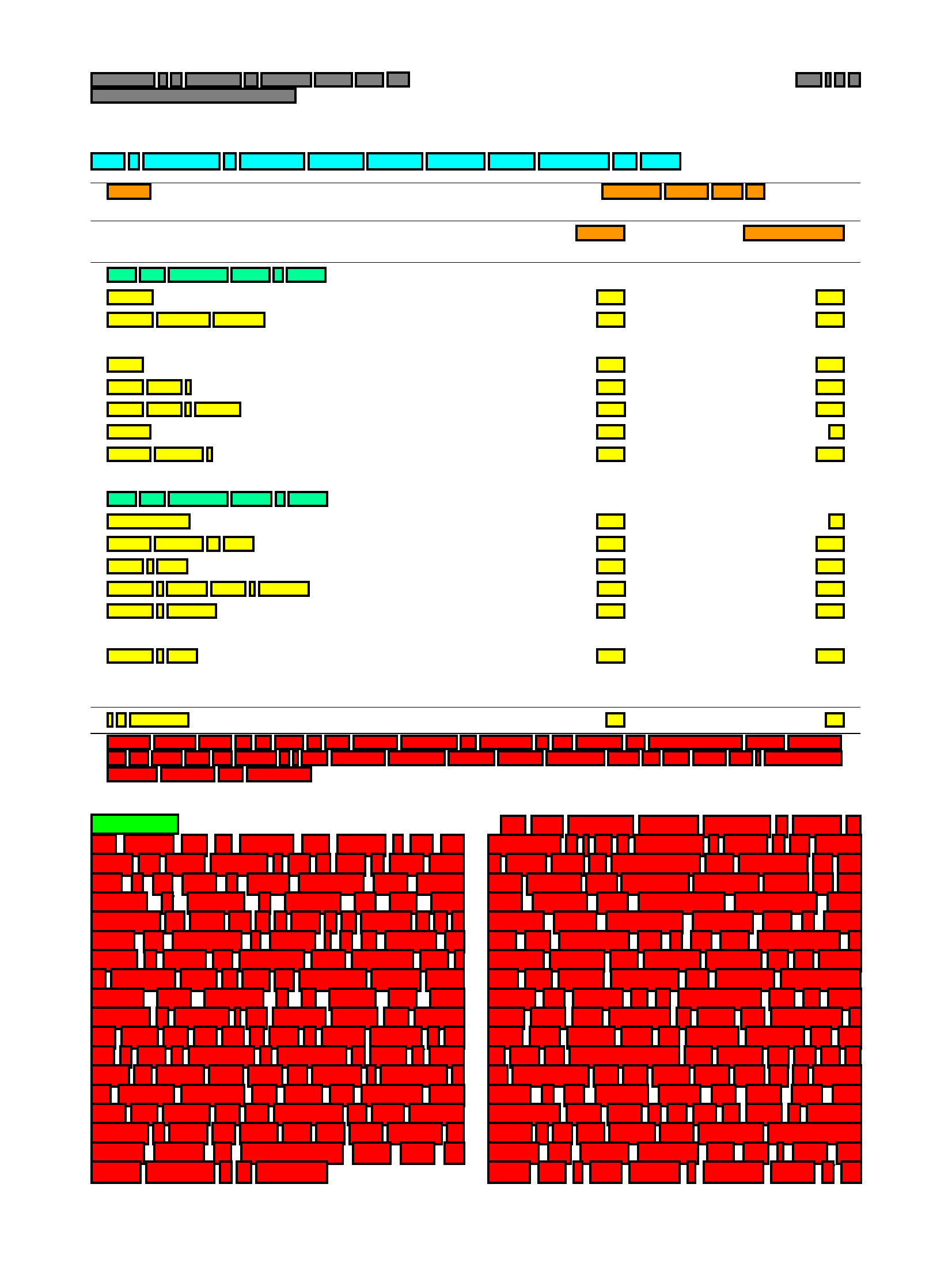}
  \caption{Spanning rows.}
  \label{fig:sfig1}
\end{subfigure}%
\begin{subfigure}{.33\textwidth}
  \centering
  \includegraphics[width=\linewidth]{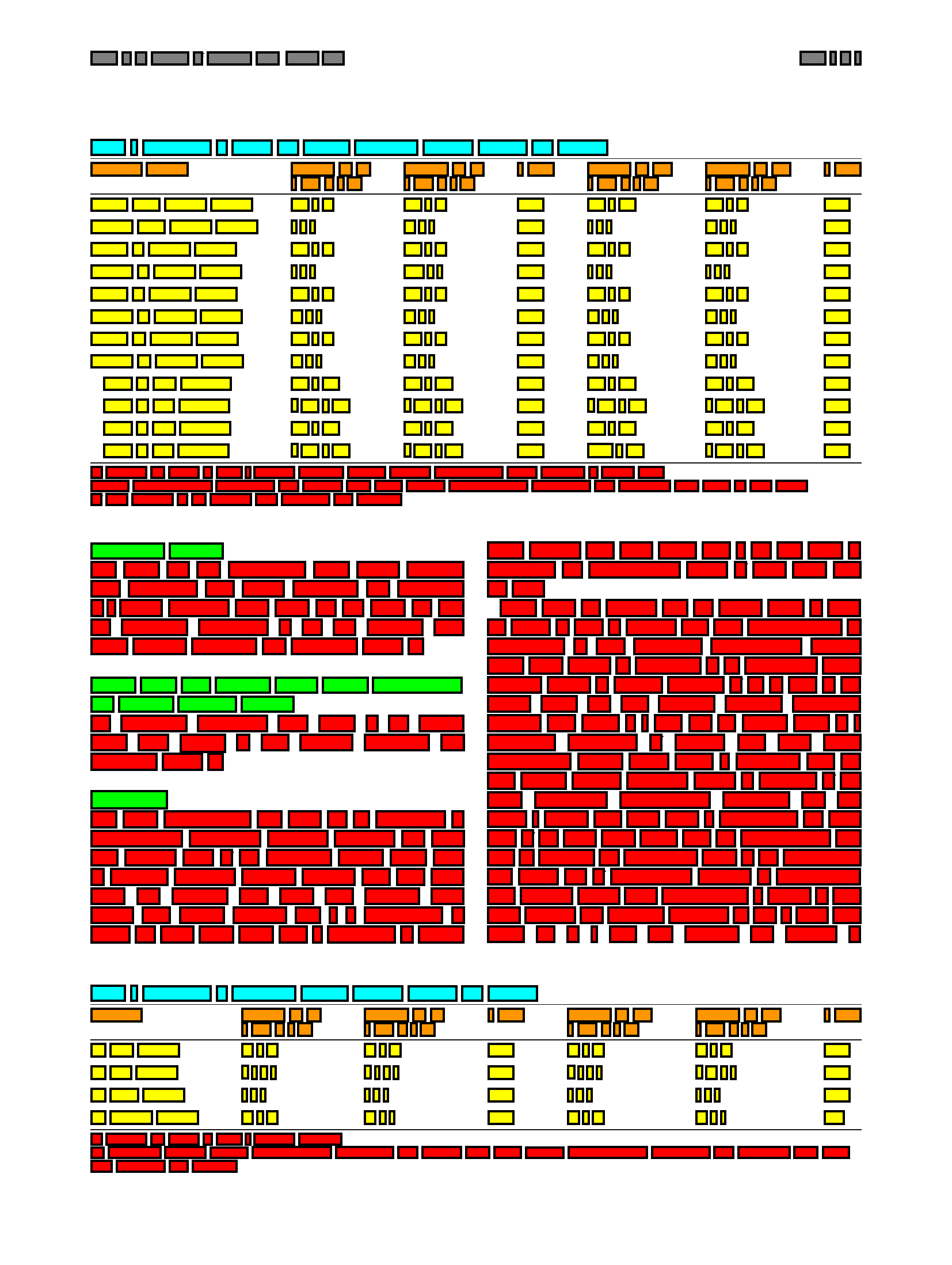}
  \caption{More out-column tables.}
  \label{fig:sfig2}
\end{subfigure}
\begin{subfigure}{.33\textwidth}
  \centering
  \includegraphics[width=\linewidth]{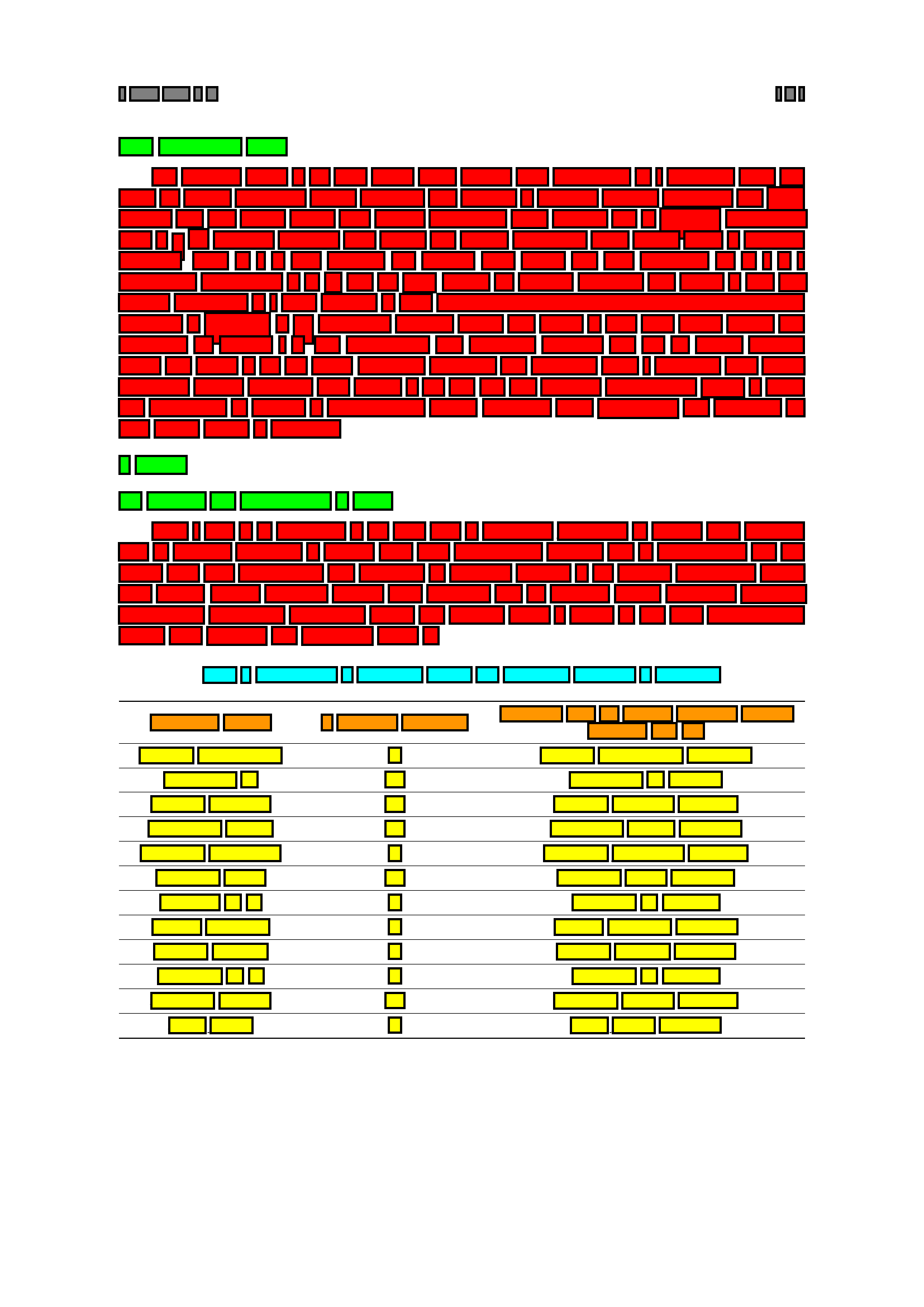}
  \caption{Single page layout.}
  \label{fig:sfig3}
\end{subfigure}
\begin{subfigure}{.33\textwidth}
  \centering
  \includegraphics[width=\linewidth]{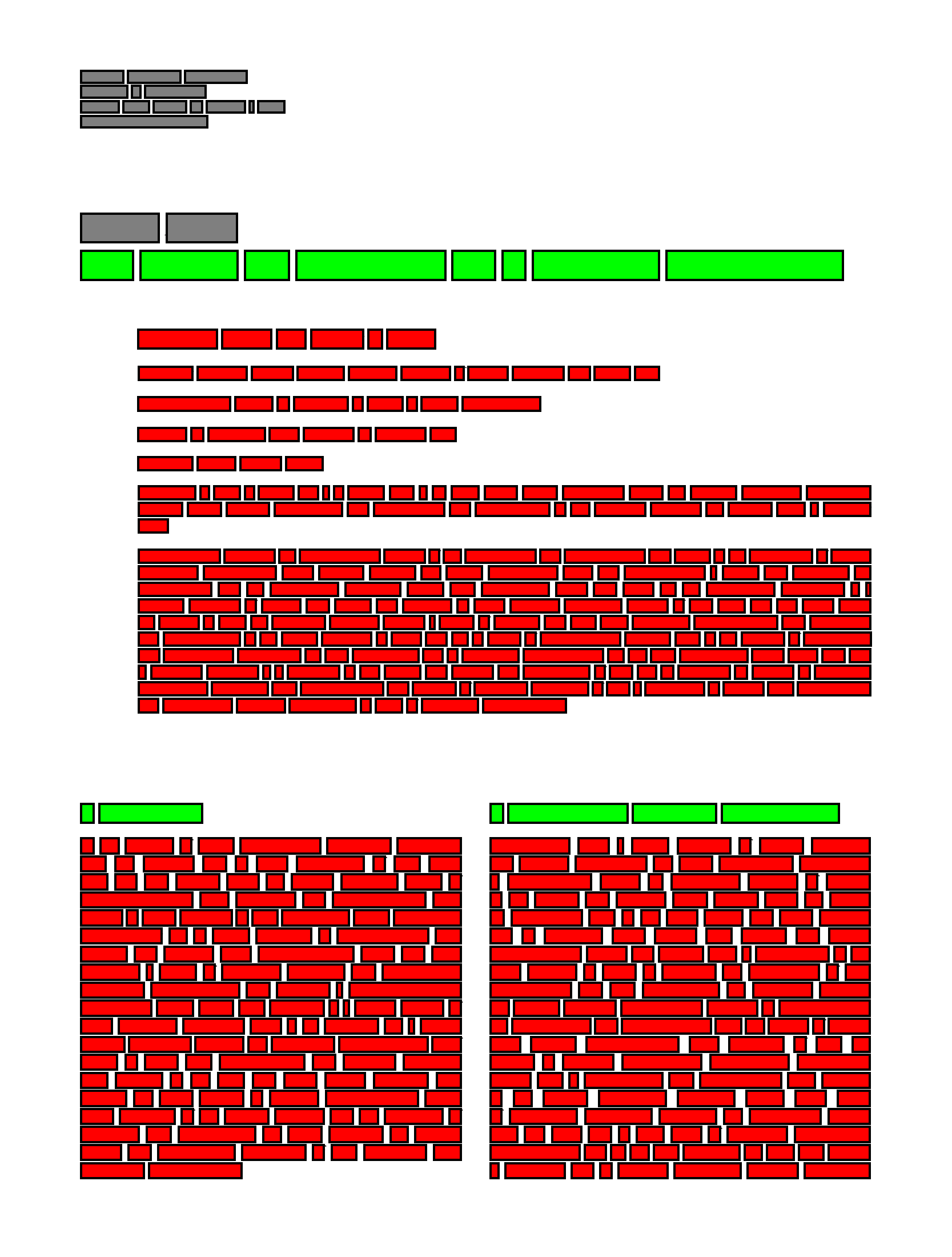}
  \caption{Title page.}
  \label{fig:sfig4}
\end{subfigure}%
\begin{subfigure}{.33\textwidth}
  \centering
  \includegraphics[width=\linewidth]{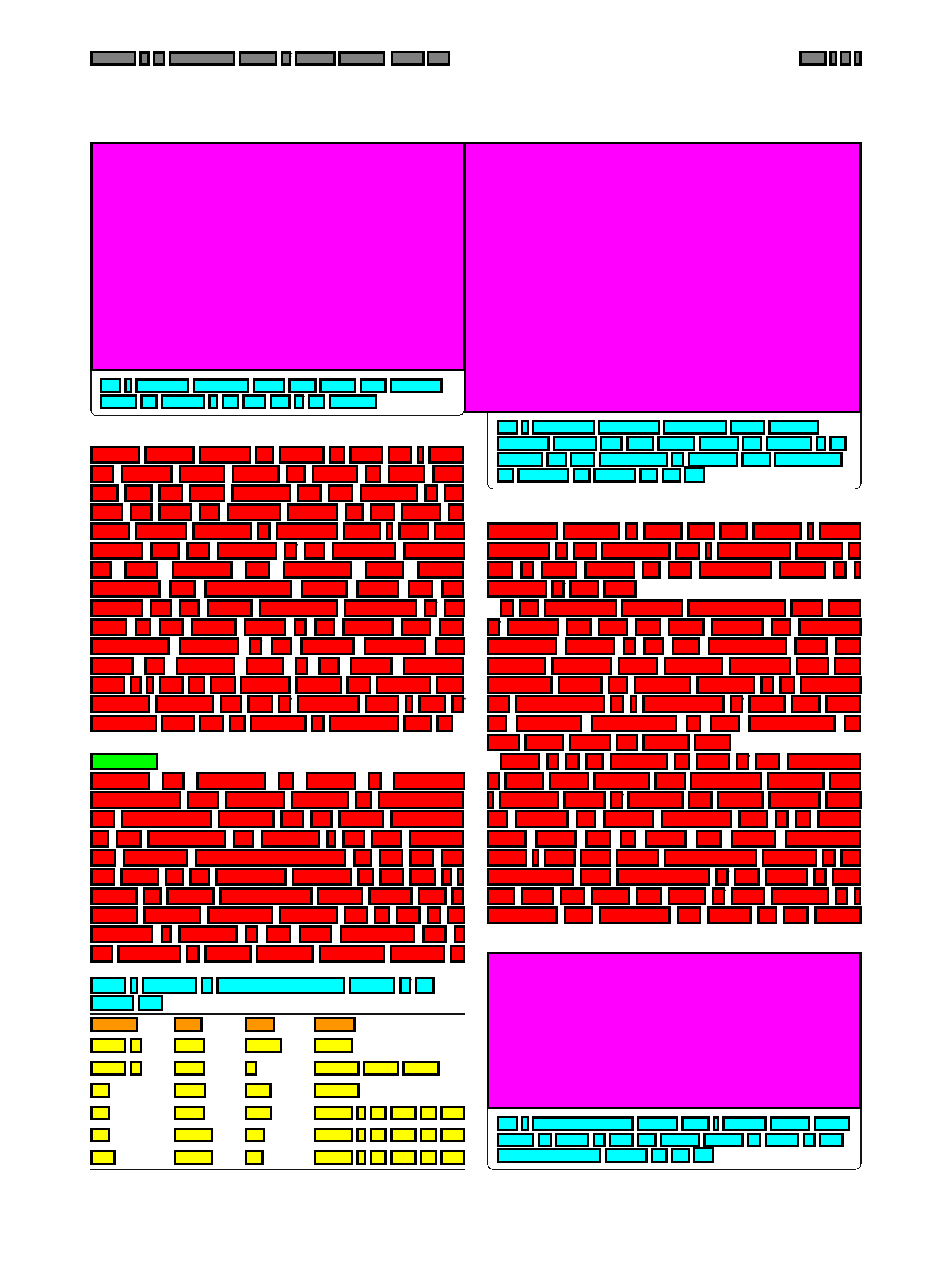}
  \caption{More images per page.}
  \label{fig:sfig5}
\end{subfigure}
\begin{subfigure}{.33\textwidth}
  \centering
  \includegraphics[width=\linewidth]{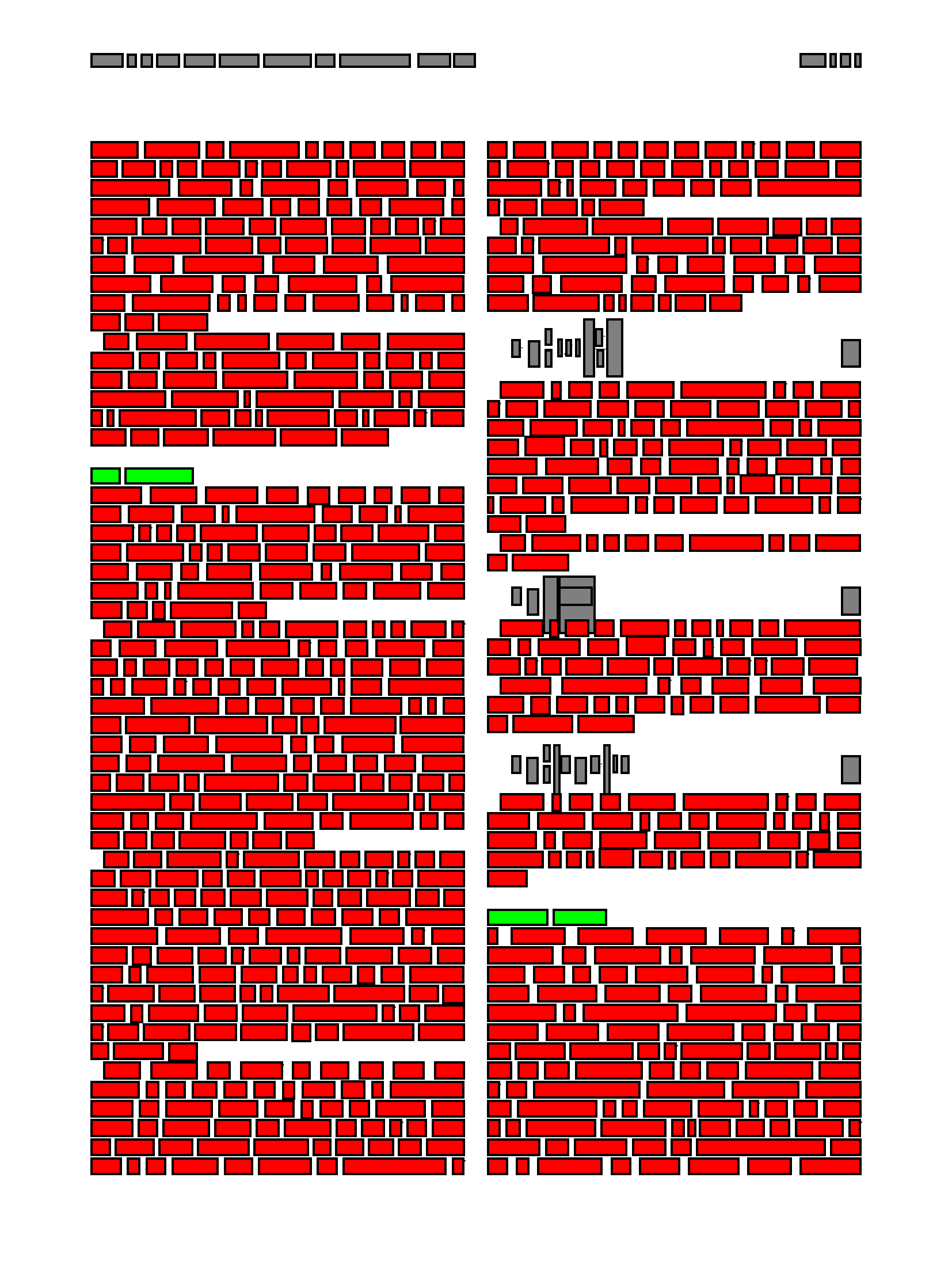}
  \caption{Formulas labeled as others.}
  \label{fig:sfig6}
\end{subfigure}
\begin{subfigure}{.33\textwidth}
  \centering
  \includegraphics[width=\linewidth]{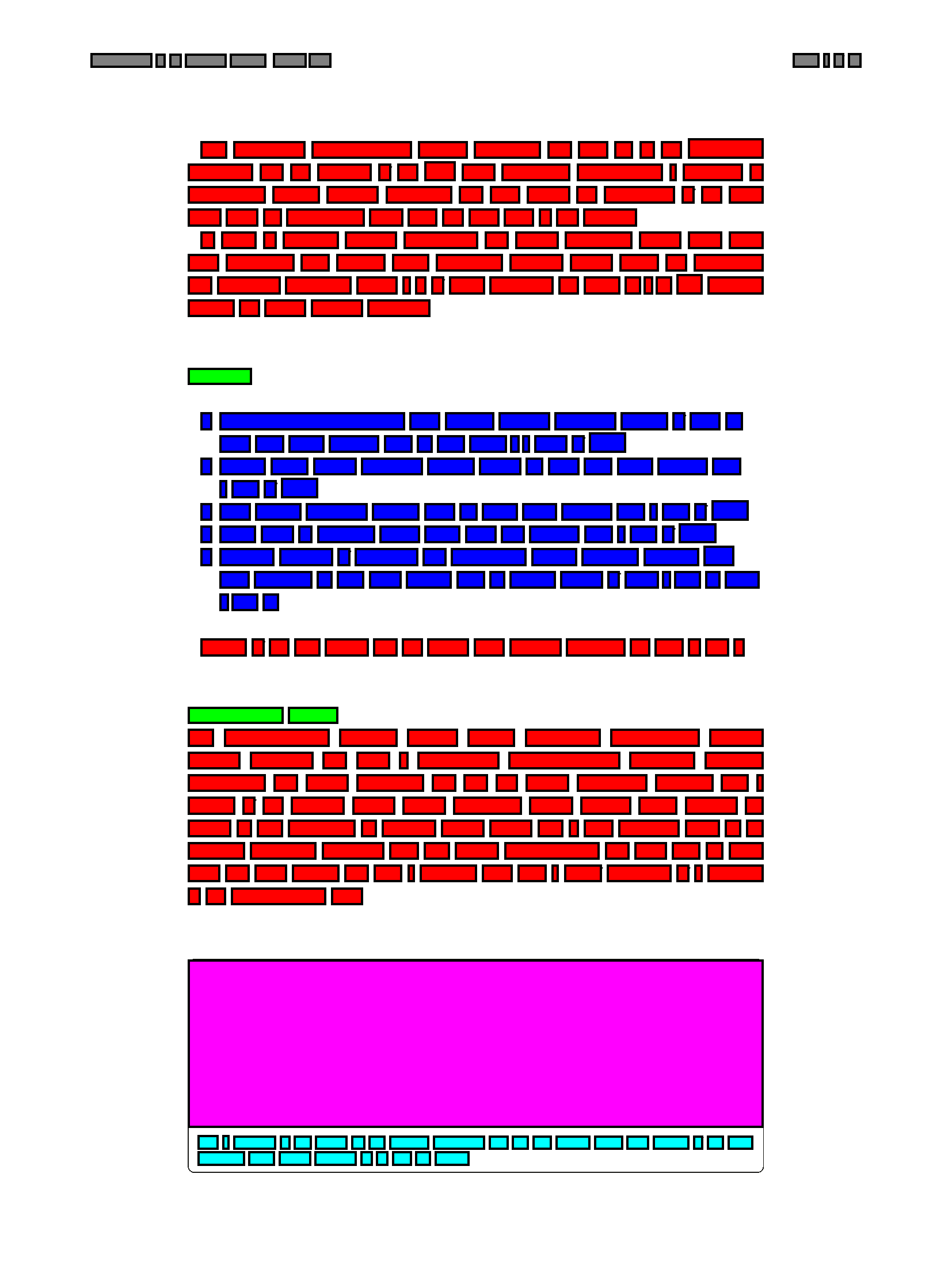}
  \caption{List example.}
  \label{fig:sfig7}
\end{subfigure}%
\begin{subfigure}{.33\textwidth}
  \centering
  \includegraphics[width=\linewidth]{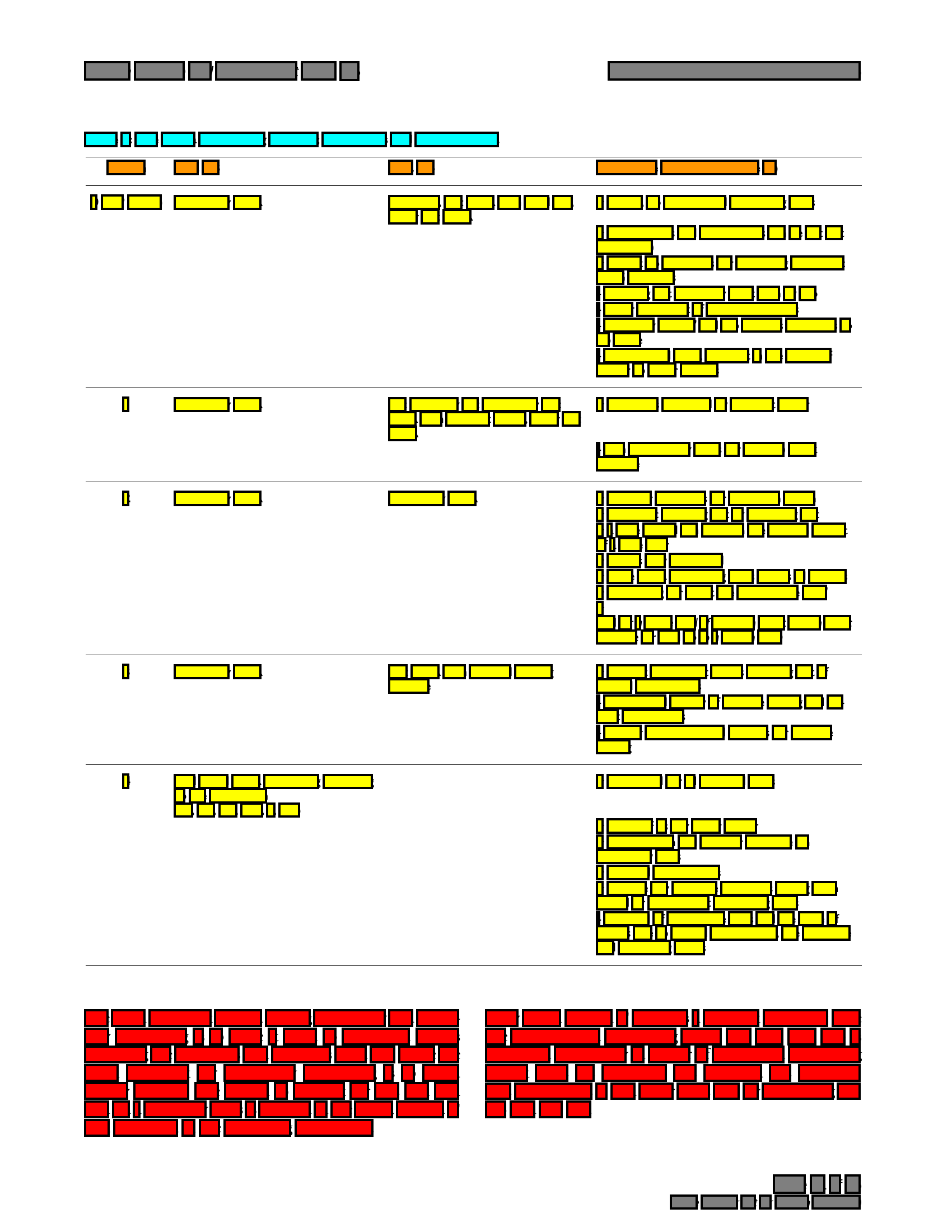}
  \caption{Full page table.}
  \label{fig:sfig8}
\end{subfigure}
\begin{subfigure}{.33\textwidth}
  \centering
  \includegraphics[width=\linewidth]{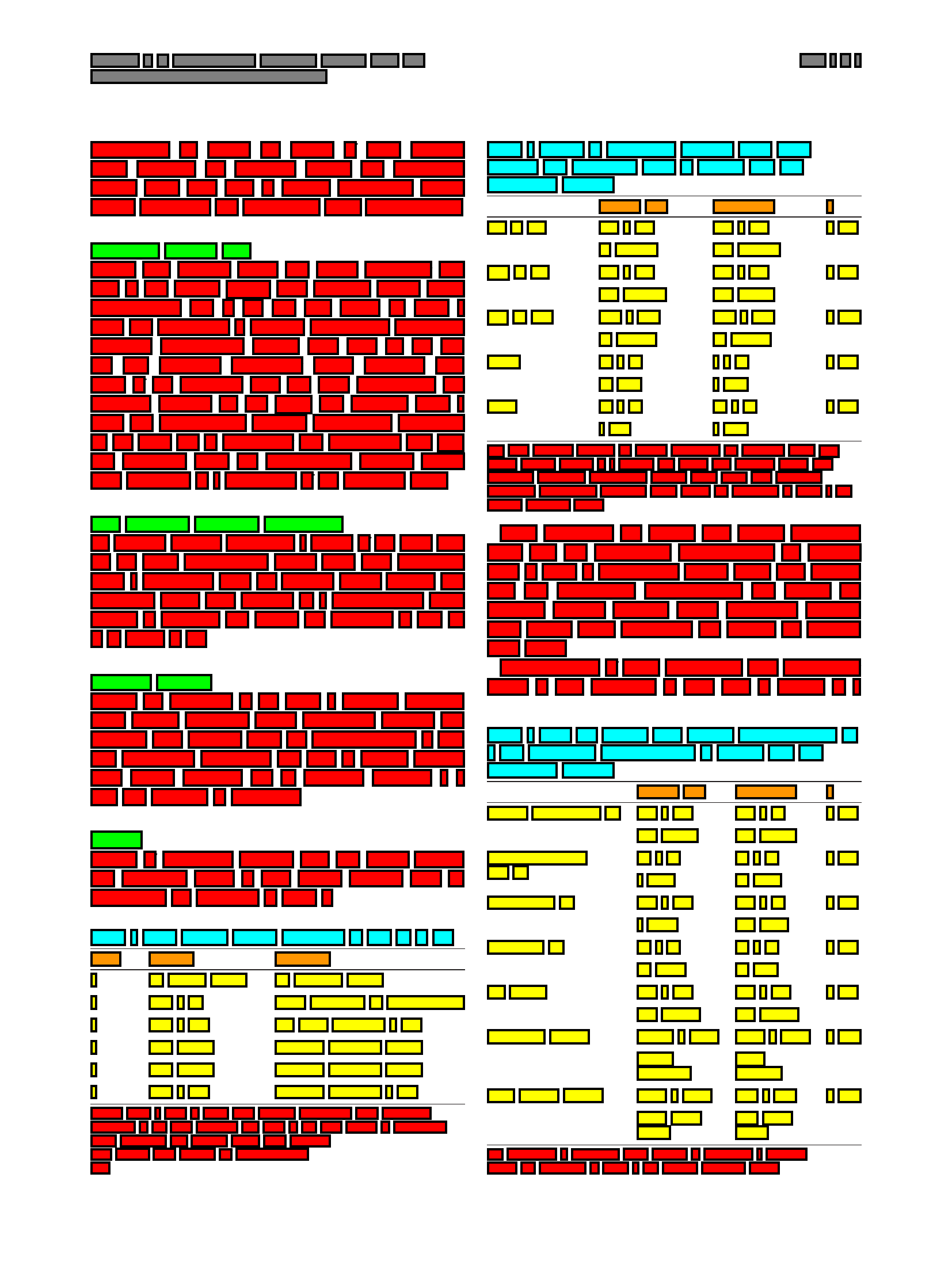}
  \caption{More in-column tables.}
  \label{fig:sfig9}
\end{subfigure}
\caption{Examples of labeled pages showing the different layouts available in the dataset. 
There can be more tables and images per page (b, i, e) either aligned or not with columns. 
There are single-column pages (c). 
As mentioned in Section \ref{sec:limitations}, there are some limitations: equations not labeled (f), missing keywords, authors, abstract information (d), and subtitles that are not distinguished from the paper title.}
\label{fig:examples}
\end{figure}
\end{document}